\documentclass[journal]{IEEEtran}
\IEEEoverridecommandlockouts

\usepackage{cite}
\usepackage{amsmath,amssymb,amsfonts}
\usepackage{algorithmic}
\usepackage{graphicx}
\usepackage{textcomp}
\usepackage{xcolor}
\usepackage{booktabs}
\usepackage{multirow}
\usepackage{array}
\usepackage[colorlinks,linkcolor=red,anchorcolor=red,citecolor=blue]{hyperref}
\usepackage[lined,boxed,ruled]{algorithm2e}

\graphicspath{{./figs/}}

\def\BibTeX{{\rm B\kern-.05em{\sc i\kern-.025em b}\kern-.08em
    T\kern-.1667em\lower.7ex\hbox{E}\kern-.125emX}}

\begin{document}

\title{MMAO-Dyn: A Metabolic Multi-Agent Optimizer for Dynamic Optimization}

\author{Jinliang Xu$^{*}$ and Liping Ma%
\thanks{Jinliang Xu received the bachelor's and master's degrees from Beijing University of Posts and Telecommunications, Beijing, China, where he also received the Ph.D. degree in computer science in 2020. He is with the China Academy of Information and Communications Technology, Beijing, China. His current research interests include heuristic optimization, agent-based applications, and blockchain systems. His e-mail: jlxufly@gmail.com and xujinliang@caict.ac.cn.}%
\thanks{Liping Ma is with the Department of Disease Control and Prevention, The Seventh Medical Center of Chinese PLA General Hospital, Beijing, China; e-mail: lipingmaqzx@163.com.}}

\maketitle

\begin{abstract}
This paper studies whether the Metabolic Multi-Agent Optimizer (MMAO) can be credibly derived into a dynamic-optimization method without replacing its core metabolic control loop by external adaptation modules. The proposed MMAO-Dyn maps private energy, communal budget, role drift, success feedback, and lifecycle turnover to a nonstationary setting in which environmental changes repeatedly invalidate previously useful local structure. We evaluate MMAO-Dyn on an 18-scenario synthetic dynamic continuous benchmark matrix covering shifted sphere, shifted Ackley, and shifted Rastrigin landscapes at $10D$, $20D$, and $30D$, with two change severities and 12 seeds per scenario. The comparison layer includes a generic MMAO variant without dynamic derivation, dynamic random search, dynamic PSO-lite, dynamic DE-lite, and three endogenous ablations. Across the full 216-run matrix, MMAO-Dyn attains mean offline error $28.07$, improving over Generic-MMAO ($29.36$), Dynamic-PSO-lite ($34.65$), Dynamic-DE-lite ($67.09$), and Dynamic-RandomSearch ($111.37$). The gains are clearest in aggregate robustness on sphere and Rastrigin families and in 10-step post-change recovery relative to the generic backbone, whereas the seed-aligned comparison with Dynamic-PSO-lite remains unfavorable in win-loss count and the \texttt{NoMemoryRefresh} ablation stays very close to the full method. The results support extending the metabolic loop to dynamic optimization, particularly through recovery-oriented resource redistribution.
\end{abstract}


\section{Introduction}
\IEEEPARstart{D}{ynamic} optimization problems differ from static optimization not only because the target moves, but because useful search behavior must be redistributed over time. A method that performs well on a fixed landscape may still fail badly when changes invalidate prior local structure, alter promising regions, or reward faster recovery over slower asymptotic precision. For this reason, dynamic optimization is a natural stress test for any framework whose identity depends on endogenous allocation of search effort rather than on one fixed move operator \cite{branke2012book,nguyen2012evolutionary,benchmark2022gmpb}.

The Metabolic Multi-Agent Optimizer (MMAO) is a cross-domain heuristic framework in which agents earn, spend, donate, and reinvest bounded resources through a private-public metabolic loop. The central hypothesis behind MMAO is not that one operator is universally best, but that useful search pressure can emerge from a closed resource economy among heterogeneous agents. This makes dynamic optimization especially relevant: if the metabolic loop really is the organizing principle of the framework, it should remain meaningful after environmental changes, when the optimizer must redirect effort instead of merely refining a static incumbent.

This paper asks whether the same metabolic logic can be derived into a credible optimizer for dynamic optimization without abandoning the core loop in favor of an externally attached change-handling module. The goal is therefore narrower than a full state-of-the-art challenge paper. We do not claim that a small synthetic study resolves the broader dynamic-optimization literature. Instead, we ask whether a recognizably derived MMAO family member can exhibit coherent post-change behavior, outperform a generic non-derived MMAO backbone, and produce interpretable evidence that the metabolic control loop remains useful under nonstationarity.

That question has value beyond this one implementation. Dynamic optimization exposes whether a framework can redirect accumulated search capital after disruption, which is a stricter test of controller coherence than steady-state static search alone.

The paper makes four contributions.
\begin{itemize}
    \item It formulates MMAO-Dyn as a dynamic derivative of MMAO in which change response, role drift, resource injection, and memory refresh are all tied to the same endogenous bookkeeping logic.
    \item It provides an implementation-level algorithmic description, including explicit state variables, update logic, and complexity considerations, so that the dynamic extension is not merely conceptual.
    \item It reports an expanded benchmark study on shifted sphere, shifted Ackley, and shifted Rastrigin scenarios with aggregate, grouped, pairwise, ablation, and recovery-window analyses.
    \item It clarifies the current empirical scope of the method: the strongest validated value lies in recovery-oriented redistribution after change, while advantages on harder multimodal cases remain mixed.
\end{itemize}

\section{Related Work}
Dynamic optimization has long been studied through evolutionary and swarm-based methods that preserve memory, maintain diversity, predict environmental drift, or partition the population into sub-swarms and subproblems \cite{branke2012book,nguyen2012evolutionary,yuan2013dynamic}. Benchmarking studies have repeatedly emphasized that dynamic optimizers should not be judged only by final-best values, but by tracking quality, offline error, and recovery after repeated changes \cite{benchmark2022gmpb}. In practice, many successful approaches rely on explicit mechanisms such as restart schedules, change detectors, archives, speciation, or transfer strategies \cite{signorelli2025perturbation,liu2023transfer}.

This literature matters to MMAO-Dyn in two ways. First, it shows that dynamic optimization typically demands more than direct reuse of a static optimizer. Second, it provides a cautionary baseline: dynamic behavior can often be obtained by attaching auxiliary modules, but such attachment does not automatically establish a coherent framework identity. MMAO variants should share an internal control logic that supports their domain-specific mechanisms.

Resource allocation is another important line of work. Adaptive budget redistribution has been studied in distributed differential evolution, cooperative particle swarm optimization, and other multi-population settings \cite{li2022distributed,liu2022cooperative}. These works show that budget allocation can affect search performance. However, their allocation rules are usually embedded in specialized optimizer architectures. MMAO-Dyn differs in that resource redistribution is not a secondary scheduling layer; it is the primary control language of the framework.

The design of MMAO-Dyn is also related to parameter-control research \cite{eiben1999parameter,karafotias2015parameter}. A common challenge in adaptive heuristics is that useful feedback is often added as a separate controller, thereby increasing algorithmic complexity and obscuring the conceptual core. Our design stance is deliberately stricter: adaptation should be explainable as a consequence of the metabolic loop itself. This principle does not eliminate all parameters, nor does it make the present implementation parameter-free, but it constrains how new dynamic mechanisms may be introduced.

Finally, the paper is informed by benchmarking methodology for derivative-free optimization under limited budgets \cite{raponi2023optimizing,stripinis2024benchmarking}. Since the current study uses a moderate synthetic matrix rather than a very large benchmark suite, it is important to present results conservatively and distinguish between a credible family-expansion result and a comprehensive performance claim.

\section{From MMAO to MMAO-Dyn}
\subsection{Design Requirement}
The design rule adopted here is intentionally strict:
\begin{quote}
all adaptive responses in MMAO-Dyn must be explainable as consequences of the metabolic resource loop, not as externally attached change-handling modules.
\end{quote}

This rule does not forbid dynamic-specific mapping. It requires that the mapping remain derivational rather than reconstructive. In other words, a mechanism is allowed only if it can be interpreted as a change in how agents earn, spend, conserve, receive, or lose resources under nonstationarity.

\subsection{State Variables}
At iteration $t$, the optimizer faces a time-varying objective $f_t(x)$. Each agent maintains
\begin{equation}
    A_i(t)=\big(x_i(t),E_i(t),\phi_i(t),m_i(t),\tau_i(t),\kappa_i(t)\big),
\end{equation}
where $x_i(t)$ is the current solution, $E_i(t)$ is private energy, $\phi_i(t)\in[0,1]$ is continuous role state, $m_i(t)$ is local memory, $\tau_i(t)$ is age or stagnation state, and $\kappa_i(t)$ summarizes recent change exposure or recovery pressure.

The population additionally maintains a communal budget $B_t$, a smoothed success statistic $\rho_t$, and a post-change pressure signal. The present implementation derives this signal from the observed jump in the incumbent quality after a regime shift:
\begin{equation}
    \Delta_t=\max\big(0, f_t(x_t^{\star})-f_{t^-}(x_{t^-}^{\star})\big),
\end{equation}
where $t^-$ denotes the iteration immediately preceding the environmental change. A normalized recovery pressure is then defined as
\begin{equation}
    r_t=\mathrm{clip}\left(\eta \frac{\Delta_t}{s_t},0,1\right),
\end{equation}
where $s_t$ is a recent gain scale and $\eta$ is a sensitivity constant. This signal determines how aggressively the method should shift toward exploration and reinvest communal resources.

\subsection{Metabolic Interpretation of Dynamic Response}
The dynamic derivation is summarized in Table~\ref{tab:mapping}. The key idea is not that environmental change simply triggers a restart, but that change alters the metabolic value of existing evidence. Old local improvements become partially stale, exploratory roles become temporarily more valuable, and communal reserves become more useful when directed toward recovery rather than toward normal maintenance.

\begin{table}[t]
\caption{Dynamic mapping of the MMAO metabolic loop.}
\label{tab:mapping}
\centering
\small
\resizebox{\columnwidth}{!}{%
\begin{tabular}{p{2.3cm}p{4.8cm}}
\toprule
MMAO concept & Dynamic reinterpretation in MMAO-Dyn \\
\midrule
Private energy & Measures current local usefulness after change; low-value agents are pruned earlier under recovery pressure. \\
Communal budget & Stores redistributable effort and is temporarily injected after change to fund scouts and recovery. \\
Role drift & Moves continuously toward exploration when previous local evidence becomes unreliable. \\
Success feedback & Rewards agents that adapt quickly in the new regime and guides later reinvestment. \\
Lifecycle turnover & Removes agents whose energy and progress stay weak after change; replacement increases search breadth. \\
Memory & Remains endogenous but is re-evaluated against the new regime rather than trusted unconditionally. \\
\bottomrule
\end{tabular}
}
\end{table}

\subsection{Operational Mechanisms}
The implemented MMAO-Dyn uses five derived mechanisms.
\begin{itemize}
    \item \textbf{Change-triggered resource injection:} after an environmental shift, the communal pool receives an additional reserve proportional to the estimated disruption.
    \item \textbf{Recovery-window reinvestment:} for a limited number of iterations after change, step sizes and role updates are biased toward exploration.
    \item \textbf{Exploration-biased role reset:} agents are not hard-restarted, but their role state is shifted toward lower-$\phi$ exploratory behavior.
    \item \textbf{Memory refresh:} archived personal-best solutions are re-evaluated under the new regime and can be replaced by refreshed local samples.
    \item \textbf{Scout injection and turnover:} a small number of new agents are spawned to widen coverage, while weak agents face stricter elimination.
\end{itemize}

All five mechanisms are expressed through the same bookkeeping language of private energy, communal budget, role state, and survival pressure. This does not make the algorithm minimal, but it does preserve a single explanatory core.

It is also useful to state explicitly what is inherited unchanged and what is only reinterpreted. The inherited part is the controller skeleton itself: bounded private energy, communal donation and spending, continuous role movement, and budget-financed turnover. What changes in the dynamic setting is the semantic meaning of success, memory reliability, and reinvestment urgency after regime shifts. These changes extend the existing controller to nonstationary conditions.

\subsection{Algorithmic Skeleton}
Algorithm~\ref{alg:mmaodyn} summarizes the optimizer at the level relevant for reproducibility. The static search operator is inherited from the continuous MMAO backbone, while dynamic behavior is introduced through reevaluation, pressure estimation, and recovery-biased redistribution.

\begin{algorithm}[t]
\caption{High-level MMAO-Dyn}
\label{alg:mmaodyn}
\DontPrintSemicolon
Initialize population, private energies, role states, communal budget, and local memories\;
Evaluate the initial regime and record the incumbent\;
\For{$t=1,\dots,T$}{
    \If{environment changes}{
        reevaluate all agents under the new regime\;
        estimate post-change gap $\Delta_t$ and recovery pressure $r_t$\;
        inject communal reserve and open a finite recovery window\;
        shift role states toward exploration, refresh memories, and add scouts\;
    }
    compute population spread, reserve target, and success statistic\;
    \ForEach{agent}{
        generate exploratory probes using role-sensitive step scales\;
        generate an exploitative candidate from personal memory and global elite\;
        reward improvement, update private energy, and donate the communal share\;
        update the continuous role state using gain, pressure, and success feedback\;
    }
    prune low-energy stagnant agents and respawn if the population falls below the minimum\;
    reinvest communal resources into promising offspring when surplus exists\;
    refine the current elite and record offline error and resource traces\;
}
\Return best trajectory and summary statistics\;
\end{algorithm}

\subsection{Complexity and Budget Interpretation}
For a population of size $N_t$ in dimension $d$, the dominant cost per iteration is objective evaluation. Each agent performs a small constant number of exploratory probes and one exploitative proposal, followed by optional elite refinement. Ignoring low-order bookkeeping terms, the algorithm therefore behaves as
\begin{equation}
    \mathcal{O}(N_t d)+\mathcal{O}(N_t C_f)
\end{equation}
per iteration, where $C_f$ is the cost of one fitness evaluation. Since the implementation uses bounded population sizes and constant-size local probing, the effective budget is controlled mainly by the iteration horizon and the population envelope rather than by an unbounded auxiliary module. This matters for fair comparison against low-budget dynamic baselines.

\section{Experimental Protocol}
\subsection{Research Questions}
The study is organized around four questions (see GitHub repository \texttt{mmao}\footnote{\url{https://github.com/wolfbrother/mmao}} or PyPI package \texttt{mmao-opt}\footnote{\url{https://pypi.org/project/mmao-opt/}}).
\begin{enumerate}
    \item \textbf{RQ1:} Can MMAO-Dyn track moving optima credibly on a small but structured dynamic benchmark matrix?
    \item \textbf{RQ2:} Does the metabolic loop generate meaningful post-change recovery behavior?
    \item \textbf{RQ3:} Is the dynamic extension recognizably derived from MMAO rather than rebuilt from scratch?
    \item \textbf{RQ4:} Which endogenous mechanisms matter most under nonstationarity?
\end{enumerate}

\subsection{Benchmark Matrix}
The benchmark matrix contains three shifted landscape families:
\begin{itemize}
    \item shifted sphere;
    \item shifted Ackley;
    \item shifted Rastrigin.
\end{itemize}

Each family is evaluated at $10D$, $20D$, and $30D$, with change severities $0.8$ and $1.2$. Change frequency is 20, 30, and 40 iterations for the three dimensions, respectively, and each run lasts 160 iterations. The seed set is fixed to 12 values per scenario, producing 216 runs per method over the full matrix.

The matrix crosses three landscape families, three dimensional regimes, and two severity levels. Its scale remains well below GMPB-scale campaigns, limiting the scope of performance comparisons.

\subsection{Compared Methods}
The baseline layer contains four comparators.
\begin{itemize}
    \item \texttt{Generic-MMAO}: the same general search backbone without the dynamic derivation.
    \item \texttt{Dynamic-RandomSearch}: a simple non-adaptive baseline that samples uniformly each iteration.
    \item \texttt{Dynamic-PSO-lite}: a lightweight dynamic particle swarm baseline with re-evaluated personal and global bests after change.
    \item \texttt{Dynamic-DE-lite}: a lightweight dynamic differential-evolution baseline with post-change reevaluation and bounded mutation-recombination search.
\end{itemize}

We additionally evaluate three ablations:
\begin{itemize}
    \item \texttt{NoChangeResponse};
    \item \texttt{NoMemoryRefresh};
    \item \texttt{FixedRoleShift}.
\end{itemize}

These ablations test how the dynamic mechanisms affect behavior and performance.

\subsection{Budget Fairness}
The comparison is budget-matched at the iteration level and approximately matched in per-iteration effort. MMAO-Dyn and Generic-MMAO use the same 160-iteration horizon, the same initialization scale, and the same dimension-dependent population envelopes: 8/10/12 initial agents and 24/28/32 maximum agents for $10D/20D/30D$. Dynamic-RandomSearch uses 16, 24, and 28 samples per iteration, while Dynamic-PSO-lite and Dynamic-DE-lite use population sizes of 14, 18, and 22, respectively. These settings place all methods in a comparable low-budget regime intended to emphasize reaction and redistribution rather than heavy offline tuning. Following the benchmarking cautions in \cite{raponi2023optimizing,stripinis2024benchmarking}, we therefore avoid presenting the study as a definitive cross-family ranking.

\subsection{Why These Baselines Are Sufficient for This Paper}
Because this is a derivation paper rather than a leaderboard paper, the baseline set is chosen to answer distinct scientific questions instead of maximizing breadth. Generic-MMAO isolates the value of the dynamic derivation itself. Dynamic-RandomSearch establishes a low-information floor. Dynamic-PSO-lite tests whether a simple swarm baseline can already cover much of the same ground without the metabolic machinery, while Dynamic-DE-lite adds a second lightweight specialist family with a different search bias. This division of roles is important. If MMAO-Dyn could not outperform Generic-MMAO, the family-expansion argument would already be weakened. If it could not beat Dynamic-RandomSearch, the dynamic extension would not be credible at all. If it compares competitively with PSO-lite and DE-lite, then the metabolic loop becomes more than a purely framework-internal story.

\subsection{Evaluation Criteria}
The main metric is offline error:
\begin{equation}
    \mathrm{OE}=\frac{1}{T}\sum_{t=1}^{T}\left(\hat f_t-f_t^\star\right),
\end{equation}
where $\hat f_t$ is the best-so-far value tracked under the active regime and $f_t^\star$ is the current optimum value. Lower values indicate better long-run tracking and recovery.

To strengthen the reading beyond a single mean, we use four complementary views.
\begin{itemize}
    \item aggregate statistics across all 216 runs;
    \item scenario-wise means and family-wise summaries;
    \item paired sign-test summaries for seed-aligned comparisons;
    \item post-change recovery windows derived from trajectory histories.
\end{itemize}

\subsection{Implementation Notes}
The benchmark is implemented in Python with deterministic seed control. Each scenario starts from a zero-shift regime and applies bounded shift updates of severity-dependent magnitude. MMAO-Dyn and Generic-MMAO share the same bounded population regime, initial energy scale, and local search backbone; the only differences arise from the dynamic-derivation switches. This is a crucial implementation choice, because it prevents the comparison with Generic-MMAO from being confounded by hidden operator changes.

\section{Results}
\subsection{Aggregate Comparison}
Table~\ref{tab:aggregate} summarizes the full 216-run aggregate results. MMAO-Dyn attains mean offline error $28.07$, improving over Generic-MMAO ($29.36$), Dynamic-PSO-lite ($34.65$), Dynamic-DE-lite ($67.09$), and Dynamic-RandomSearch ($111.37$). The improvement over Generic-MMAO is modest in absolute size and supports the value of the dynamic extension. The gains over Dynamic-RandomSearch and Dynamic-DE-lite are larger. The comparison with Dynamic-PSO-lite is favorable in aggregate mean but needs finer interpretation, because the paired comparisons remain unfavorable even though the tougher scenarios pull the overall average in MMAO-Dyn's favor.

\begin{table}[t]
\caption{Aggregate offline-error summary across the full dynamic matrix. Lower is better.}
\label{tab:aggregate}
\centering
\small
\resizebox{\columnwidth}{!}{%
\begin{tabular}{lccccc}
\toprule
Method & Runs & Mean & Std & Median & Best \\
\midrule
MMAO-Dyn & 216 & 28.07 & 39.37 & 4.23 & 0.20 \\
Generic-MMAO & 216 & 29.36 & 40.21 & 6.06 & 0.23 \\
Dynamic-PSO-lite & 216 & 34.65 & 55.03 & 2.55 & 0.31 \\
Dynamic-DE-lite & 216 & 67.09 & 96.79 & 16.42 & 1.83 \\
Dynamic-RandomSearch & 216 & 111.37 & 122.18 & 78.80 & 6.91 \\
\bottomrule
\end{tabular}
}
\end{table}

\begin{figure}[t]
\centering
\includegraphics[width=\columnwidth]{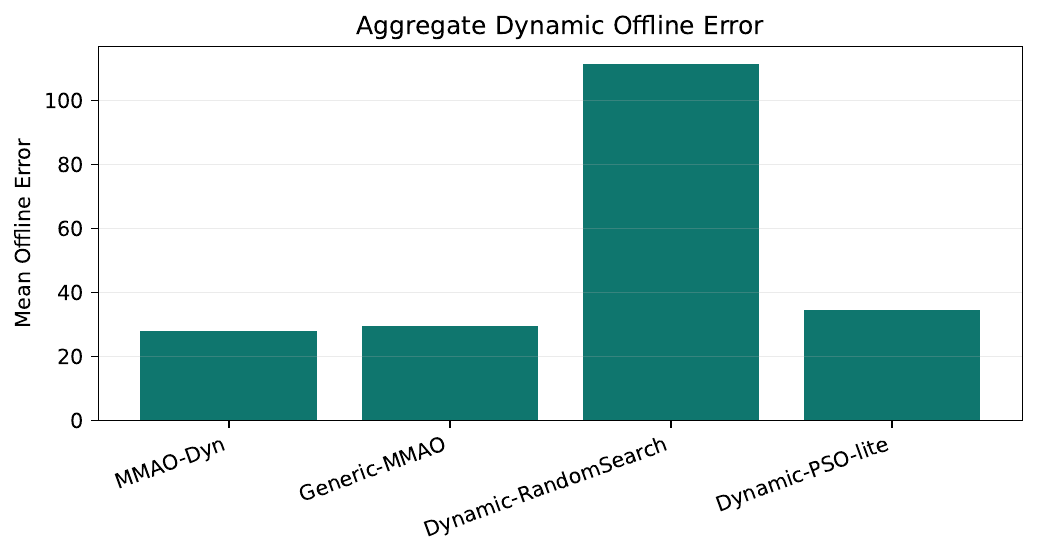}
\caption{Aggregate mean offline error on the expanded 18-scenario matrix. MMAO-Dyn improves over the generic backbone and the two weaker external baselines, while the comparison with PSO-lite requires a more nuanced paired reading.}
\label{fig:aggregate-bar}
\end{figure}

\subsection{Pairwise and Rank Reading}
The aggregate mean alone is not sufficient for dynamic optimization, because scenario-specific variance can be large. Table~\ref{tab:pairwise} therefore reports seed-aligned pairwise comparisons. Against Generic-MMAO, MMAO-Dyn wins 168 of 216 comparisons and loses 48, with a two-sided sign-test $p=8.44\times 10^{-17}$. Against Dynamic-RandomSearch, it wins all 216 comparisons. Against Dynamic-PSO-lite, however, the pairwise picture is explicitly unfavorable: MMAO-Dyn wins 65 and loses 151, even though its aggregate mean is better. This occurs because PSO-lite is extremely strong on the smoother sphere and Ackley cases, whereas MMAO-Dyn is more robust on the harder tails of the matrix, especially Rastrigin and higher dimensions. The scenario-average rank summary therefore reads differently from the aggregate mean: Dynamic-PSO-lite attains average rank 1.67, MMAO-Dyn 1.78, Generic-MMAO 2.56, and Dynamic-RandomSearch 4.00.

\begin{table}[t]
\caption{Pairwise seed-aligned comparison of MMAO-Dyn against the three baseline comparators included in the formal sign-test layer. MeanDiff is $\mathrm{OE}_{\text{MMAO-Dyn}}-\mathrm{OE}_{\text{baseline}}$.}
\label{tab:pairwise}
\centering
\small
\resizebox{\columnwidth}{!}{%
\begin{tabular}{lcccccc}
\toprule
Comparator & Wins & Losses & Ties & MeanDiff & MedianDiff & Sign-test $p$ \\
\midrule
Generic-MMAO & 168 & 48 & 0 & -1.29 & -0.71 & $8.44\times10^{-17}$ \\
Dynamic-PSO-lite & 65 & 151 & 0 & -6.58 & 1.07 & $4.62\times10^{-9}$ \\
Dynamic-RandomSearch & 216 & 0 & 0 & -83.30 & -69.09 & $1.90\times10^{-65}$ \\
\bottomrule
\end{tabular}
}
\end{table}

\subsection{Scenario Structure}
The expanded matrix hides a clear structural split. Table~\ref{tab:family} groups results by landscape family. On shifted sphere and shifted Ackley, MMAO-Dyn remains stronger than Generic-MMAO and much stronger than Dynamic-RandomSearch and Dynamic-DE-lite, but Dynamic-PSO-lite is especially strong. On shifted Rastrigin, by contrast, MMAO-Dyn improves over Generic-MMAO and clearly improves over both external lightweight baselines in family mean. The dimension-wise grouped reading is also instructive: MMAO-Dyn is slightly better than Dynamic-PSO-lite at $10D$, nearly tied with Generic-MMAO at $20D$, and clearly stronger than both at $30D$.

\begin{table}[t]
\caption{Mean offline error grouped by landscape family. Lower is better.}
\label{tab:family}
\centering
\small
\resizebox{\columnwidth}{!}{%
\begin{tabular}{lccccc}
\toprule
Landscape family & MMAO-Dyn & Generic-MMAO & Dynamic-PSO-lite & Dynamic-DE-lite & Dynamic-RandomSearch \\
\midrule
Sphere & 4.94 & 7.12 & 2.35 & 18.86 & 81.63 \\
Ackley & 3.19 & 3.66 & 1.88 & 3.92 & 8.05 \\
Rastrigin & 76.09 & 77.29 & 99.73 & 178.49 & 244.43 \\
\bottomrule
\end{tabular}
}
\end{table}

This split is important for interpreting the value of the dynamic derivation. On smooth and moderately rugged landscapes, the main benefit is fast recovery relative to a generic non-derived MMAO backbone; however, a well-behaved swarm baseline can still be very strong. On the multimodal Rastrigin family, the dynamic derivation becomes more valuable because robustness after change matters more than immediate smooth tracking. The benefit of the dynamic extension therefore varies by landscape family.

\subsection{What the Scenario Split Suggests Mechanistically}
The sphere, Ackley, and Rastrigin families stress different parts of the dynamic logic. Sphere primarily tests how quickly the optimizer can redirect effort after a coherent global shift. Ackley adds a middle regime in which the landscape is still globally structured but less trivially smooth. In these settings, the recovery-window idea is naturally valuable: stale local evidence should be discounted, search breadth should rise briefly, and communal resources should be spent aggressively on reacquisition. Rastrigin, by contrast, tests whether that same redistribution can separate useful broadening from merely diffusing effort across many deceptive basins. The mixed $20D$ behavior and the paired losses to PSO-lite therefore do not simply indicate ``failure''; they identify the current mechanism's main bottleneck. MMAO-Dyn already reacts, but it does not yet discriminate post-change structure sharply enough on every rugged or easier-smooth regime.

\subsection{Ablation Analysis}
The ablations help answer whether the dynamic derivation is genuinely consequential. Removing explicit post-change response raises aggregate mean offline error from $28.07$ to $29.20$. Removing memory refresh yields $28.03$, and fixing role-shift behavior yields $28.26$. The aggregate differences are therefore not dramatic, and in one case an ablation is numerically slightly better. Their interpretation becomes clearer after grouping by family. Table~\ref{tab:ablationfamily} shows that on sphere and Ackley the degradation from removing change response is visible, whereas on Rastrigin the relative order is much tighter and the present memory-refresh design is not yet uniformly advantageous.

\begin{table}[t]
\caption{Family-wise ablation summary. Lower is better.}
\label{tab:ablationfamily}
\centering
\small
\resizebox{\columnwidth}{!}{%
\begin{tabular}{lcccc}
\toprule
Landscape family & MMAO-Dyn & NoChangeResponse & NoMemoryRefresh & FixedRoleShift \\
\midrule
Sphere & 4.94 & 7.12 & 4.92 & 4.94 \\
Ackley & 3.19 & 3.71 & 3.11 & 3.16 \\
Rastrigin & 76.09 & 76.76 & 76.07 & 76.69 \\
\bottomrule
\end{tabular}
}
\end{table}

This ablation pattern supports a restrained but meaningful claim. The strongest current evidence does not say that every endogenous mechanism is indispensable on every landscape. Rather, it says that the change-response part of the metabolic loop matters most on those scenarios where post-change redistribution is the main difficulty, while the present memory-refresh and role-shift realizations remain plausible but not yet fully optimized.

\section{Recovery Diagnostics}
\subsection{Trajectory-Level Interpretation}
\begin{figure}[t]
\centering
\includegraphics[width=\columnwidth]{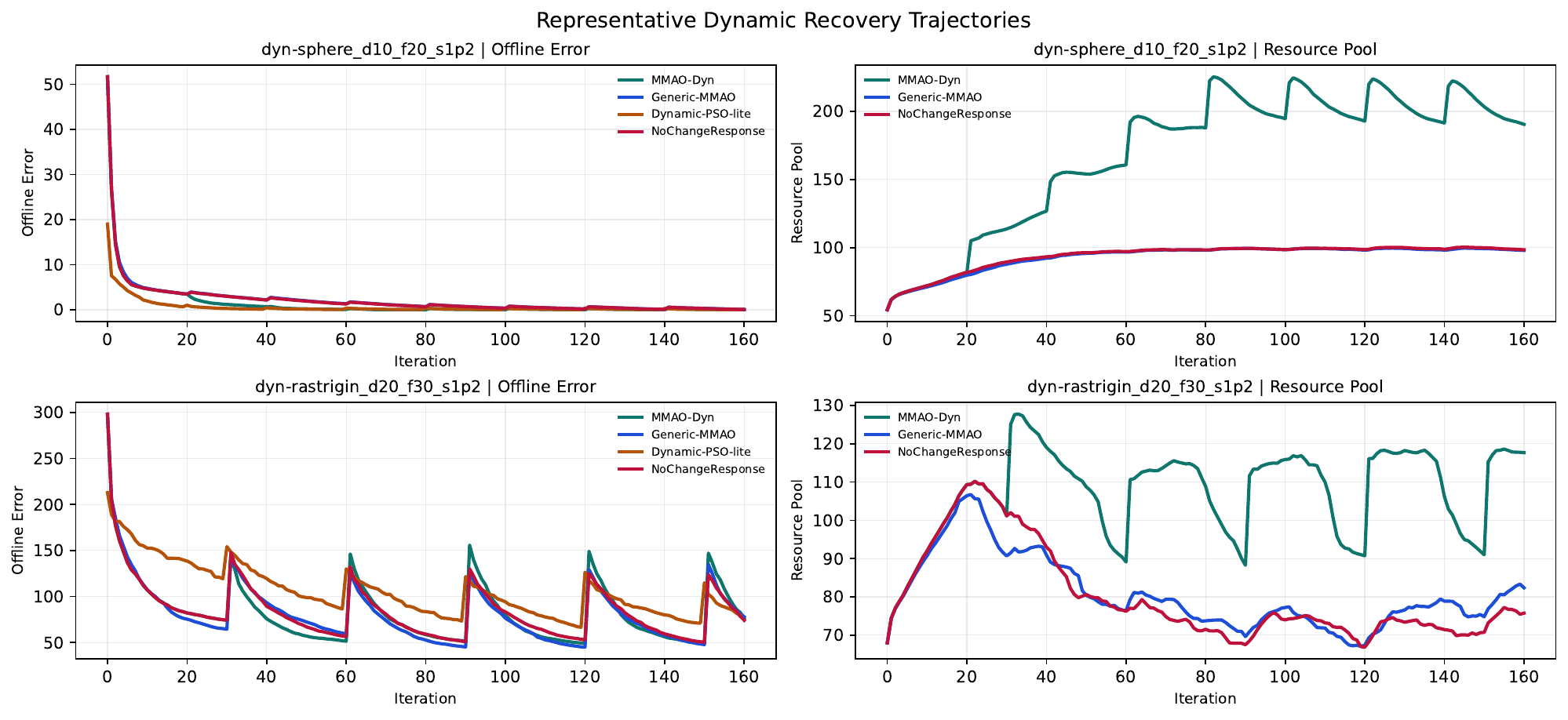}
\caption{Representative recovery trajectories. The left panels track offline error and communal resource evolution on a shifted sphere scenario; the right panels show the same on a shifted Rastrigin scenario.}
\label{fig:recovery}
\end{figure}

Offline-error tables alone are not enough for a dynamic family-expansion paper. Figure~\ref{fig:recovery} therefore shows representative trajectories for a sphere and a Rastrigin scenario. On the shifted sphere case, MMAO-Dyn exhibits clearer post-change drops in offline error than Generic-MMAO and \texttt{NoChangeResponse}. The communal resource trace also becomes more active around change points, which is consistent with the intended interpretation of recovery-window reinvestment.

On the shifted Rastrigin case, the same mechanism is still visible, but the relationship between communal redistribution and score improvement becomes noisier. Multimodality can limit the benefit of broader exploration after change. The dynamic extension is therefore interpretable, but not universally sufficient.

\subsection{Recovery-Window Statistics}
To quantify this interpretation, Table~\ref{tab:recoverywindow} reports post-change window statistics derived from the recorded histories. The first two columns average offline error over the 5-iteration and 10-iteration windows immediately following a detected environmental shift. The third column reports the mean jump at change onset. MMAO-Dyn is not the method with the smallest immediate jump; Dynamic-PSO-lite and even \texttt{NoChangeResponse} can look smoother at change onset on easier scenarios. However, MMAO-Dyn improves the 10-iteration recovery average relative to Generic-MMAO and Dynamic-RandomSearch and remains close to Dynamic-PSO-lite, which is more closely aligned with the paper's metabolic redistribution claim.

\begin{table}[t]
\caption{Aggregate recovery-window statistics across all detected changes. Lower is better.}
\label{tab:recoverywindow}
\centering
\small
\resizebox{\columnwidth}{!}{%
\begin{tabular}{lccc}
\toprule
Method & Mean post-change OE (5 iters) & Mean post-change OE (10 iters) & Mean jump at change \\
\midrule
MMAO-Dyn & 33.59 & 29.02 & 22.32 \\
Generic-MMAO & 33.51 & 30.02 & 20.06 \\
Dynamic-PSO-lite & 31.08 & 28.95 & 10.30 \\
Dynamic-DE-lite & 51.79 & 50.82 & 2.03 \\
Dynamic-RandomSearch & 104.82 & 100.61 & 25.96 \\
\bottomrule
\end{tabular}
}
\end{table}

These window averages sharpen the paper's main interpretation. First, MMAO-Dyn improves over Generic-MMAO on the 10-step recovery view ($29.02$ versus $30.02$), which is the most direct test of recovery-oriented redistribution after change. Second, the gap to Dynamic-PSO-lite is very small on the same view ($29.02$ versus $28.95$), indicating that MMAO-Dyn is competitive in sustained short-horizon recovery even when it loses many seed-aligned cases. Third, the \texttt{NoMemoryRefresh} ablation slightly improves the 10-step value to $28.81$, confirming that the current derivation is credible but not yet submechanism-optimal. This directly supports the paper's main conclusion: the present dynamic derivation is most convincing as a coherent recovery logic, while several internal choices still expose room for refinement.

\subsection{A Failure-Mode Reading}
One useful way to read the negative cases is to ask what kind of mistake the algorithm is making. The evidence suggests that MMAO-Dyn is not primarily suffering from inertia. After change, it does respond. Nor is it suffering from total instability, since the aggregate means remain competitive and the method stays well ahead of random search and DE-lite. The more likely issue is \emph{over-broad recovery}: resource injection and exploration-biased role drift can re-open the search effectively, but on smoother cases they may surrender too much immediate local quality to methods such as PSO-lite, while on rugged landscapes they may still be insufficiently selective about where the reopened effort should go. This suggests that future improvements should strengthen how the loop identifies structurally promising post-change directions.

\section{Discussion}
\subsection{What the Study Establishes}
The strongest conclusion of this paper is not that MMAO-Dyn dominates the dynamic-optimization literature. It is that the MMAO framework can be extended into a new problem family while preserving a recognizable internal identity. The dynamic method is not a trivial relabeling of a static optimizer: it improves on Generic-MMAO in aggregate, wins decisively in paired comparisons against that non-derived backbone, stays clearly ahead of the weaker external baselines, and shows trajectory-level behavior that matches the intended metabolic interpretation.

\subsection{Why This Still Counts as an Independent Paper}
A dynamic extension requires a distinct scientific question, a domain-specific mapping, and empirical evidence beyond the original cross-domain MMAO formulation \cite{xu2026mmao}. The paper's central question is about dynamic derivation, not general static capability. Its main evidence comes from change-sensitive metrics and recovery windows that are irrelevant in a static study. And its main conclusion is specific: the metabolic loop appears most useful as a recovery-oriented redistribution principle under change.

Seen this way, dynamic optimization is not merely another benchmark family. It is a sharper framework test because environmental change directly asks whether accumulated search capital can be redirected rather than merely spent harder. A controller that only works by steady-state refinement can look adequate on static tasks yet fail here. The value of MMAO-Dyn is therefore that it tests whether the resource loop can survive disruption while preserving a coherent internal identity.

\subsection{Why the Results Are Structurally Interesting}
The results are also informative beyond this particular implementation. They suggest that resource-centered frameworks may have a natural advantage in settings where the main question is how to reallocate effort after disruption. This is consistent with related work on adaptive resource allocation \cite{li2022distributed,liu2022cooperative}, but MMAO-Dyn differs in that the allocation mechanism is not added as a separate scheduling layer. Instead, it is woven directly into agent survival, reproduction, role drift, and reserve use.

At the same time, the study also reveals the current boundary of that idea. Recovery-oriented redistribution is not by itself a complete answer to dynamic search across all regimes. The smooth-family paired losses to PSO-lite and the very small gap between MMAO-Dyn and \texttt{NoMemoryRefresh} indicate that one can validate the framework-level dynamic derivation without proving that every current mechanism is already the best possible realization. These results identify recovery-oriented redistribution as the best-supported part of the current extension.

\subsection{Threats to Validity}
Several limitations should be kept explicit.
\begin{itemize}
    \item The benchmark family is synthetic and much smaller than GMPB-scale studies, even after the present expansion to 18 scenarios and 216 runs.
    \item The baseline layer is intentionally lightweight rather than specialist-heavy.
    \item The dynamic problems are continuous only; no dynamic combinatorial results are included here.
    \item The statistical layer is deliberately simple and oriented toward paired directional evidence rather than exhaustive multiple-comparison analysis.
    \item The implementation still contains several hyperparameters, even though their interactions are metabolically structured.
\end{itemize}

These limits do not invalidate the present family-expansion claim, but they do narrow the empirical scope. The paper demonstrates derivational viability and interpretable recovery value, not comprehensive superiority over the broader dynamic-optimization state of the art.

\subsection{Design Lessons for Future MMAO Variants}
The present study also suggests broader lessons for the MMAO family. First, the framework seems most coherent when every adaptation can be written in the same resource language. Second, trajectory diagnostics are especially valuable for such a framework, because aggregate means alone can hide whether the metabolic story is actually visible in behavior. Third, family expansion should be selective: not every new domain requires a large number of extra mechanisms, but every mechanism that is introduced should be metabolically interpretable. In that sense, MMAO-Dyn is both a derived optimizer and a methodological test of how the family should evolve.

The study also sharpens the parameter-light boundary for future dynamic variants. If additional change-handling coefficients are introduced freely, the framework quickly collapses back into a collection of manually tuned schedules. The more principled direction is to keep dynamic sensitivity concentrated in a small number of state-derived quantities such as disruption pressure, memory reliability, and communal reserve availability, so that adaptation still looks like a consequence of the loop rather than an attached supervisory layer.

\subsection{Future Work}
There are four especially natural next steps.
\begin{itemize}
    \item Scale the study to larger dynamic suites such as generalized moving-peaks-style benchmarks.
    \item Strengthen the baseline layer with more specialized uninformed and informed dynamic optimizers.
    \item Extend the same metabolic logic to dynamic combinatorial settings, where memory, repair, and reassignment pressures interact differently.
    \item Push the framework closer to stronger endogenous adaptation by reducing the number of manually exposed dynamic coefficients, in line with parameter-control principles \cite{eiben1999parameter,karafotias2015parameter}, while re-deriving memory refresh and role drift more selectively from the metabolic loop.
\end{itemize}

\section{Conclusion}
This paper presented MMAO-Dyn, a dynamic optimizer derived from the MMAO metabolic loop for nonstationary continuous landscapes. On an expanded 18-scenario, 216-run synthetic benchmark matrix, MMAO-Dyn improves over Generic-MMAO and Dynamic-RandomSearch clearly, outperforms Dynamic-DE-lite comfortably, and achieves a better aggregate mean than Dynamic-PSO-lite despite losing many seed-aligned pairwise comparisons to that smoother-case specialist. The comparison with Dynamic-PSO-lite, the closeness of \texttt{NoMemoryRefresh}, and the mixed behavior on some intermediate settings mean that the paper does not claim universal dominance or fully mature submechanisms. The main value of the study is therefore methodological as much as empirical: it shows that MMAO can generate a recognizable family member in a new domain without being rebuilt from scratch, and it identifies recovery-oriented resource redistribution as the most convincing current mechanism to strengthen in future dynamic variants.

The paper should therefore be read as evidence of derivational viability in a demanding nonstationary setting, not as a completed answer to the broader dynamic-optimization benchmark landscape.

\section*{Acknowledgement}
The related publications in Table~\ref{tab:mmao-research-series} provide complementary views of the MMAO research series.

\begin{table}[!ht]
\centering
\caption{Other publications in the MMAO research series.}
\label{tab:mmao-research-series}
\footnotesize
\setlength{\tabcolsep}{5pt}
\renewcommand{\arraystretch}{1.12}
\begin{tabular}{@{}p{\columnwidth}@{}}
\hline
\textbf{Related publication} \\
\hline
\emph{MMAO: A Metabolic Multi-Agent Optimizer with Endogenous Resource Allocation for Continuous and Discrete Optimization}~\cite{xu2026mmao}.\newline
\href{https://arxiv.org/abs/2606.28109}{arXiv:2606.28109}
\\[5pt]
\emph{Minimal MMAO: A Resource-Closed-Loop Framework for Adaptive Metaheuristic Search}~\cite{xu2026minimalmmao}.\newline
\href{https://arxiv.org/abs/2606.30450}{arXiv:2606.30450}
\\[5pt]
\emph{A Large-Scale Empirical Evaluation of MMAO Under Fair-Budget Continuous and Discrete Benchmarks}~\cite{xu2026largescale}.\newline
\href{https://arxiv.org/abs/2606.31584}{arXiv:2606.31584}
\\[5pt]
\emph{MMAO-Cls: Metabolic Multi-Agent Optimization for Joint Feature Selection and Classifier Tuning}~\cite{xu2026mmaocls}.\newline
\href{https://arxiv.org/abs/2607.01539}{arXiv:2607.01539}
\\[5pt]
\emph{Mechanism and Stability Analysis of Metabolic Closed-Loop Metaheuristics}~\cite{xu2026mechanism}.\newline
\href{https://arxiv.org/abs/2607.01551}{arXiv:2607.01551}
\\[5pt]
\emph{MMAO: A Unified Resource-Regulated Controller for Cross-Domain Optimization}~\cite{xu2026unifiedmmao}.\newline
\href{https://doi.org/10.1109/ACCESS.2026.3752083}{IEEE Access, 2026}
\\[5pt]
\hline
\end{tabular}
\end{table}

\bibliographystyle{IEEEtran}
\bibliography{Ref}

\end{document}